\documentclass[11pt,letterpaper]{article}
\usepackage{lipsum}
\usepackage{emnlp2017}
\usepackage{times}
\usepackage{latexsym}
\usepackage{bm}
\usepackage{graphicx}
\usepackage{mathtools}
\usepackage{amsfonts}
\usepackage{amsmath}
\usepackage{enumitem}
\usepackage{float}
\floatstyle{plaintop}
\restylefloat{table}
\usepackage[tableposition=top]{caption}
\emnlpfinalcopy

\def\emnlppaperid{1161}

\newcommand*{\affaddr}[1]{#1} % No op here. Customize it for different styles.
\newcommand*{\affmark}[1][*]{\textsuperscript{#1}}
\newcommand*{\affemail}[1]{\texttt{#1}}

\title{Learning Translations via Matrix Completion}

\author{%
	Derry Wijaya\affmark[1], Brendan Callahan\affmark[1], John Hewitt\affmark[1], Jie Gao\affmark[1], Xiao Ling\affmark[1], \\
	\textbf{Marianna Apidianaki}\affmark[12] and \textbf{Chris Callison-Burch}\affmark[1]\\
	\affaddr{\affmark[1]Computer and Information Science Department, University of Pennsylvania}\\
	\affaddr{\affmark[2]LIMSI, CNRS, Universit\'{e} Paris-Saclay, 91403 Orsay}\\	
	\affemail{derry@seas.upenn.edu}
}

\date{}
\hypersetup{draft}
\begin{document}
\maketitle
\begin{abstract}
	Bilingual Lexicon Induction is the task of learning  word  translations  without  bilingual parallel corpora. We model this task as a matrix completion problem, and present an effective and extendable framework for completing the matrix. This method harnesses diverse bilingual and monolingual signals, each of which may be incomplete or noisy. Our model achieves state-of-the-art performance for both high and low resource languages.
\end{abstract}

\section{Introduction}
Machine translation (MT) models typically require large, sentence-aligned bilingual texts to learn good translation models \citep{wu2016google, sennrich2015improving, koehn2003statistical}. However, for many language pairs, such parallel texts may only be available in limited quantities, which is problematic. Alignments at the word- or subword- levels \citep{sennrich2015neural} can be inaccurate in the limited parallel texts, which can in turn lead to inaccurate translations. Due to the low quantity and thus coverage of the texts, there may still be ``out-of-vocabulary" words encountered at run-time. The Bilingual Lexicon Induction (BLI) task \citep{rapp1995identifying}, which learns word translations from monolingual or comparable corpora, is an attempt to alleviate this problem. The goal is to use plentiful, more easily obtainable, monolingual or comparable data to infer word translations  and reduce the need for parallel data to learn good translation models. The word translations obtained by BLI can, for example, be used to augment MT systems and improve alignment accuracy, coverage, and translation quality \citep{gulcehre2016pointing,callison2006improved,daume11lexicaladapt}. 

\begin{figure}
	\includegraphics[width=\linewidth]{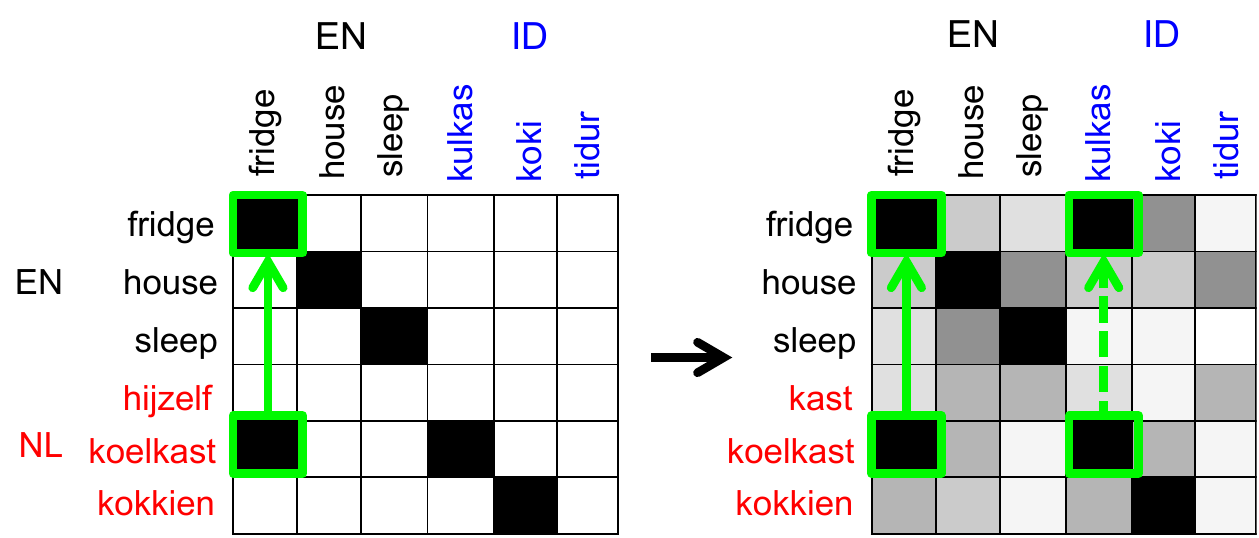}
%	\caption{Indonesian to English translation can benefit from projected translations of loan words from Dutch. Here, darker shades indicate higher confidence that the corresponding words in the row are translations of the words in the column.}
\caption{Our framework allows us to use a diverse range of signals to learn translations, including incomplete bilingual dictionaries, information from related languages (like Indonesian loan words from Dutch shown here), word embeddings, and even visual similarity cues. }
	\label{fig:relatedlanguage}
\end{figure}

Previous research has explored different sources for estimating translation equivalence from monolingual corpora  \citep{schafer2002inducing,klementiev2006weakly,irvine2013supervised,irvine2017comprehensive}. These monolingual signals, when combined in a supervised model, can enhance end-to-end MT for low resource languages \citep{klementiev2012toward,irvine2016end}. More recently, similarities between words in different languages have been approximated by constructing a shared bilingual word embedding space with different forms of bilingual supervision \citep{bicompare:16}.

We present a framework for learning translations by combining diverse  signals of translation that are each potentially sparse or noisy. We use matrix factorization (MF), which has been shown to be effective for harnessing incomplete or noisy distant supervision from multiple sources of information \cite{fan2014distant, rocktaschel2015injecting}. MF is also shown to result in good cross-lingual representations for tasks such as alignment \cite{goutte2004aligning}, QA \cite{zhou2013statistical}, and cross-lingual word embeddings \cite{shi2015learning}. 

Specifically, we represent translation as a matrix with source words in the columns and target words in the rows, and model the task of learning translations as a matrix completion problem. Starting from some observed translations (e.g., from existing bilingual dictionaries,) we infer missing translations in the matrix using MF with a Bayesian Personalized Ranking (BPR) objective \cite{rendle2009bpr}. We select BPR for a number of reasons: (1) BPR has been shown to outperform traditional supervised methods in the presence of positive-only data \cite{riedel2013relation}, which is true in our case since we only observe positive translations. (2) BPR is easily extendable to incorporate additional signals for inferring missing values in the matrix \cite{he2016vbpr}. Since observed translations may be sparse, i.e. the ``cold start" problem in the matrix completion task, incorporating additional signals of translation equivalence estimated on monolingual corpora is useful. (3) BPR is also shown to be effective for multilingual transfer learning  \cite{verga2016multilingual}. For low resource source languages, there may be related, higher resource languages from which we can project available translations (e.g., translations of loan words) to the target language (Figure \ref{fig:relatedlanguage}). 

We conduct large scale experiments to learn translations from both low and high resource languages to English and achieve state-of-the-art performance on these languages. Our main contributions are as follows:
\begin{itemize} [noitemsep,topsep=0pt]
	\item We introduce a MF framework that learns translations by integrating diverse bilingual and monolingual signals of translation, each potentially noisy/incomplete.
	\item The framework is easily extendable to incorporate additional signals of translation equivalence. Since ours is a framework for integration, each signal can be improved separately to improve the overall system. 
	\item Large scale experiments on both low and high resource languages show the effectiveness of our model, outperforming the current state-of-the-art.
	\item We make our code, datasets, and output translations publicly available.\footnote{http://www.cis.upenn.edu/\%7Ederry/translations.html}  
\end{itemize}

\section{Related Work}
\paragraph{Bilingual Lexicon Induction}
Previous research has used different sources for estimating translations from monolingual corpora. Signals such as contextual, temporal, topical, and ortographic similarities between words are used to measure their translation equivalence \citep{schafer2002inducing,klementiev2006weakly,irvine2013supervised,irvine2017comprehensive}.  

With the increasing popularity of word embeddings, many recent works approximate similarities between words in different languages by constructing a shared bilingual embedding space \cite{klementiev2012inducing,zou2013bilingual,vulic2013study,mikolov2013exploiting,faruqui2014improving,ap2014autoencoder,gouws2014bilbowa,luong2015bilingual,lu2015deep,bicompare:16}. In the shared space, words from different languages are represented in a language-independent manner such that similar words, regardless of language, have similar representations. Similarities between  words can then be measured in the shared space. One approach to induce this shared space is to learn a mapping function between the languages' monolingual semantic spaces \cite{mikolov2013exploiting,dinu2014improving}. The mapping relies on seed translations which can be from existing dictionaries or be reliably chosen from pseudo-bilingual corpora of comparable texts e.g., Wikipedia with interlanguage links. \citet{vulic2015bilingual} show that by learning a linear function with a reliably chosen seed lexicon, they outperform other models with more expensive bilingual signals for training on benchmark data. 

Most prior work on BLI however, either makes use of only one monolingual signal or uses unsupervised methods (e.g., rank combination) to aggregate the signals. \citet{irvine2016end} show that combining monolingual signals in a supervised logistic regression model produces higher accuracy word translations than unsupervised models. More recently, \citet{vulic2016multi} show that their multi-modal model that employs a simple weighted-sum of word embeddings and visual similarities can improve translation accuracy. These works show that there is a need for combining diverse, multi-modal monolingual signals of translations. In this paper, we take this step further by combining the monolingual signals with bilingual signals of translations from existing bilingual dictionaries of related, ``third" languages.

\paragraph{Bayesian Personalized Ranking (BPR)} 
Our approach is based on extensions to the probabilistic model of MF in collaborative filtering \cite{koren2009matrix,rendle2009bpr}. We represent our translation task as a matrix with source words in the columns and target words in the rows (Figure \ref{fig:relatedlanguage}). Based on some observed translations in the matrix found in a seed dictionary, our model learns low-dimensional feature vectors that encode the latent properties of the words in the row and the words in the column. The dot product of these vectors, which indicate how ``aligned" the source and the target word properties are, captures how likely they are to be translations. 

Since we do not observe false translations in the seed dictionary, the training data in the matrix consists only of positive translations. The absence of values in the matrix does not imply that the corresponding words are not translations. In fact, we seek to predict which of these missing values are true. %BPR approach to MF \cite{rendle2009bpr} is effective for learning from these positive-only data, unlike traditional supervised models that may be sensitive to the choice of negative data \cite{riedel2013relation}. 
The BPR approach to MF \cite{rendle2009bpr} formulates the task of predicting missing values as a ranking task. With the assumption that observed true translations should be given higher values than unobserved translations, BPR learns to optimize the difference between values assigned to the observed translations and values assigned to the unobserved translations. 

However, due to the sparsity of existing bilingual dictionaries (for some language pairs such dictionaries may not exist), the traditional formulation of MF with BPR suffers from the ``cold start" issue \cite{gantner2010learning,he2016vbpr,verga2016multilingual}. In our case, these are situations in which some source words have no translations to any word in the target or related languages. For these words, additional information, e.g., monolingual signals of translation equivalence or language-independent representations such as visual representations, must be used. 

We use bilingual translations from the source to the target language, English, obtained from Wikipedia page titles with interlanguage links. Since Wikipedia pages in the source language may be linked to pages in languages other than English, we also use high accuracy, crowdsourced translations \cite{pavlick2014language} from these \textit{third} languages to English as additional bilingual translations. To alleviate the cold start issue, when a source word has no existing known translation to English or other third languages, our model backs-off to additional signals of translation equivalence estimated based on its word embedding and visual representations. 

\section{Method}
In this section, we describe our framework for integrating bilingual and monolingual signals for learning translations. First we formulate the task of Bilingual Lexicon Induction, and introduce our model for learning translations given observed translations and additional monolingual/language-independent signals. Then we derive our learning procedure using the BPR objective function.

\paragraph{Problem Formulation}
Given a set of source words $F$, a set of target words $E$, the pair $\langle e$, $f \rangle$ where $e \in E$ and $f \in F$ is a candidate translation with an associated score $x_{e,f} \in [0,1]$ indicating the confidence of the translation. The input to our model is a set of observed translations $T$ := $\{\langle e, f\rangle\ |\ x_{e,f} = 1\}$.  These could come from an incomplete bilingual dictionary. We also add word identities to the matrix i.e., we define $T^{identity} := \{\langle e, e\rangle\}$, where $T^{identity} \subset T$. The task of Bilingual Lexicon Induction is then to generate \textit{missing} translations: for a given source word $f$ and a set of target words $\{e\ |\ \langle e, f\rangle \notin T\}$, predict the score ${x}_{e,f}$ of how likely it is for $e$ to be a translation of $f$.

\paragraph{Bilingual Signals for Translation}
One way to predict ${x}_{e,f}$ is by using matrix factorization. The problem of predicting ${x}_{e,f}$ can be seen as a task of estimating a matrix $X$ : $E \times F$. $X$ is approximated by a matrix product of two low-rank matrices $P : |E| \times k$ and $Q : |F| \times k$: 
$$
\hat{X} := P Q^T
$$ 
where $k$ is the rank of the approximation. Each row $p_e$ in $P$ can be seen as a feature vector describing the latent properties of the target word $e$, and each row $q_f$ of $Q$ describes the latent properties of the source word $f$. Their dot product encodes how aligned the latent properties are and, since these vectors are trained on observed translations, it encodes how likely they are to be translation of each other. Thus, we can write this formulation of predicted scores $\hat{x}_{e,f}$ with MF as:
\begin{equation} \label{eq:1}
\hat{x}_{e,f}^{\mathrm{MF}} = p_e^T q_f = \sum^{k}_{i=1} p_{ei}\ .\ q_{fi}
\end{equation}

\paragraph{Auxiliary Signals for Translation}
\begin{figure}
	\centering
	\includegraphics[scale=0.35]{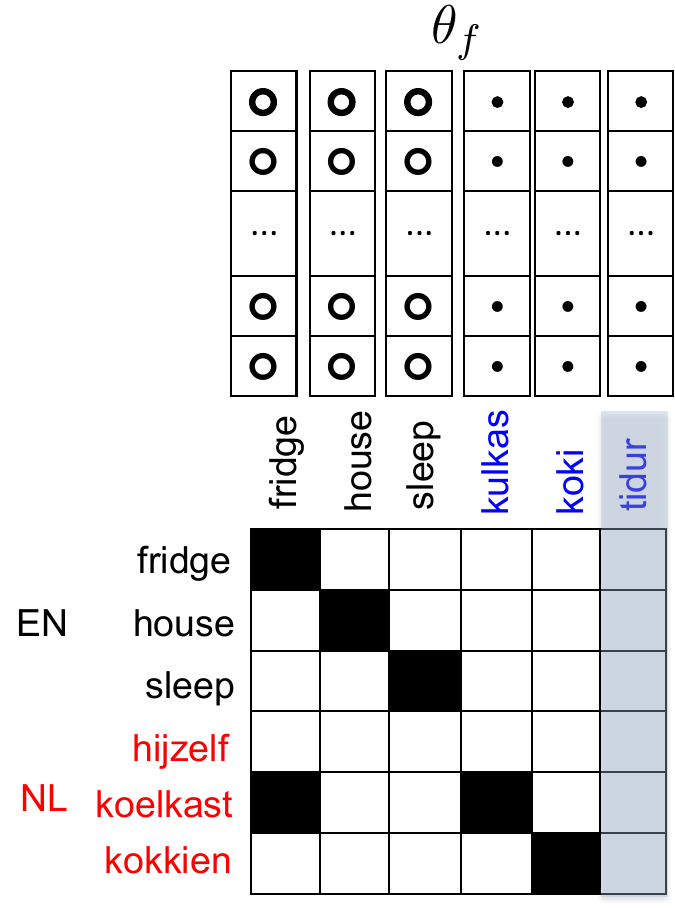}
	\caption{The word \textit{tidur} (id) is a cold word with no associated translation in the matrix. Auxiliary features $\theta_f$ about the words can be used to predict translations for cold words.}
	\label{fig:cold}
\end{figure}
Because the observed bilingual translations may be sparse, the MF approach can suffer from the existence of \textit{cold} items: words that have none or too few associated observed translations to estimate their latent dimensions accurately (Figure \ref{fig:cold}). Additional signals for measuring translation equivalence can alleviate this problem. Hence, in the case of cold words, we use a formulation of $\hat{x}_{e,f}$ that involves auxiliary features about the words in the predicted $\hat{x}_{e,f}$:
\begin{equation}\label{eq:2}
\hat{x}_{e,f}^{\mathrm{AUX}} = \theta_e^T \theta_f + \beta^T \theta_f
\end{equation}
$\theta_f$ represents an auxiliary information about the cold word $f$ e.g., its word embedding or visual features. $\theta_e$ is a feature vector to be trained, whose dot product with $\theta_f$ models the extent to which the word $e$ matches the auxiliary features of word $f$. In practice, learning $\theta_e$ amounts to learning a classifier, one for each target word $e$ that learns weights $\theta_e$ given the feature vectors $\theta_f$ of its translations. $\beta$ models the targets' overall bias toward a given word $f$. 
%\begin{figure}
%	\centering
%	\includegraphics[scale=0.35]{auxiliary.pdf}
%	\caption{Auxiliary features $\theta_i$ about the words can be used %to predict translations for cold words.}
%	\label{fig:aux}
%\end{figure}

Since each word can have multiple additional feature vectors, we can formulate $\hat{x}_{e,f}^{\mathrm{AUX}}$ as a weighted sum of available auxiliary features\footnote{We omit bias terms for brevity.}:
$$
\hat{x}_{e,f}^{\mathrm{AUX}} = \alpha_1\ \theta_e^T \theta_f + \alpha_2\ \gamma_e^T \gamma_f + ... + \alpha_n\ \delta_e^T \delta_f
$$
where $\alpha_m$ are parameters assigned to control the contribution of each auxiliary feature. 

In practice, we can combine the MF and auxiliary formulations by defining:
$$
\hat{x}_{e,f} = \hat{x}_{e,f}^{\mathrm{MF}} + \hat{x}_{e,f}^{\mathrm{AUX}}
$$
However, since bilingual signals that are input to $\hat{x}_{e,f}^{\mathrm{MF}}$ are often precise but sparse, while monolingual signals that are input to $\hat{x}_{e,f}^{\mathrm{AUX}}$ are often noisy and not sparse, in our model we only back-off to the less precise $\hat{x}_{e,f}^{\mathrm{AUX}}$ for cold source words that have none or too few associated translations (more details are given in the experiments, Section~\ref{experiments}). For other source words, we use $\hat{x}_{e,f}^{\mathrm{MF}}$ to predict. 

\paragraph{Learning with Bayesian Personalized Ranking}
Unlike traditional supervised models that try to maximize the scores assigned to positive instances (in our case, observed translations), the objective of Bayesian Personalized Ranking (BPR) is to maximize the \textit{difference} in scores assigned to the observed translations compared to those assigned to the unobserved translations. Given a training set $D$ consisting of triples of the form $\langle e, f, g \rangle$, where $\langle e, f \rangle \in T$ and $\langle e, g \rangle \notin T$, BPR wants to maximize $\hat{x}_{e,f,g}$, defined as:
$$
\hat{x}_{e,f,g}= \hat{x}_{e,f} - \hat{x}_{e,g}
$$
where $\hat{x}_{e,f}$ and $\hat{x}_{e,g}$ can be defined either by eq. \ref{eq:1} or eq. \ref{eq:2} (for cold words). Specifically, BPR optimizes \cite{rendle2009bpr}:
$$
\sum_{\langle e, f, g \rangle \in D} \mathrm{ln}\ \sigma (\hat{x}_{e,f,g}) - \lambda_{\Theta} ||\Theta||^2
$$
where $\sigma$ is the logistic sigmoid function, $\Theta$ is the parameter vector of $\hat{x}_{e,f,g}$ to be trained, and $\lambda_{\Theta}$ is its hyperparameter vector. BPR can be trained using stochastic gradient ascent where  a triple $\langle e, f, g \rangle$ is sampled from $D$ and parameter updates are performed:
$$
\Theta \leftarrow \Theta + \eta . (\sigma (-\hat{x}_{e,f,g}) \frac{\partial \hat{x}_{e,f,g}}{\partial \Theta} - \lambda_{\Theta}\Theta)
$$
$\eta$ is the learning rate. Hence, for the MF formulation of $\hat{x}_{e,f,g}$, we can sample a triple $\langle e, f, g \rangle$ from $D$ and update its parameters as: 
\begin{align*}
p_e &\leftarrow p_e + \eta . (\sigma (-\hat{x}_{e,f,g}^{\mathrm{MF}})(q_f - q_g) - \lambda_{P}\ p_e) \\
q_f &\leftarrow q_f + \eta . (\sigma (-\hat{x}_{e,f,g}^{\mathrm{MF}})(p_e) - \lambda_{Q^+}\ q_f)\\
q_g &\leftarrow q_g + \eta . (\sigma (-\hat{x}_{e,f,g}^{\mathrm{MF}})(-p_e) - \lambda_{Q^-}\ q_g)
\end{align*}
while for the auxiliary formulation of $\hat{x}_{e,f,g}$, we can sample a triple $\langle e, f, g \rangle$ from $D$ and update its parameters as:
\begin{align*}
\theta_e &\leftarrow \theta_e + \eta . (\sigma (-\hat{x}_{e,f,g}^{\mathrm{AUX}})(\theta_f - \theta_g) - \lambda_{\Theta}\ \theta_e) \\
\beta &\leftarrow \beta + \eta . (\sigma (-\hat{x}_{e,f,g}^{\mathrm{AUX}})(\theta_f - \theta_g) - \lambda_{\beta}\ \beta)
\end{align*}
\section{Experiments} \label{experiments}
To implement our approach, we extend the implementation of BPR in \textsc{Librec}\footnote{https://www.librec.net/index.html} which is a publicly available Java library for recommender systems. 

We evaluate our model for the task of Bilingual Lexicon Induction (BLI). Given a source word $f$, the task is to rank all candidate target words $e$ by their predicted translation scores $\hat{x}_{e,f}$. We conduct large-scale experiments on 27 low- and high-resource source languages and evaluate their translations to English. We use the 100K most frequent words from English Wikipedia as candidate English target words ($E$). 

At test time, for each source language, we evaluate the top-10 accuracy ($Acc_{10}$): the percent of source language words in the test set for which a correct English translation appears in the top-10 ranked English candidates. 

\subsection{Data} 
\subsubsection{Test sets}
We use  benchmark  test  sets  for  the  task  of  bilingual  lexicon induction to  evaluate  the  performance  of  our  model. The \textbf{\textsc{VULIC1000}} dataset \cite{vulic2016bilingual} comprises 1000 nouns in Spanish, Italian, and Dutch, along with their one-to-one ground-truth word translations in English. 

We construct a new test set (\textbf{\textsc{CrowdTest}}) for a larger set of 27 languages from crowdsourced dictionaries \cite{pavlick2014language}. %To make them comparable to \textbf{\textsc{VULIC1000}}, 
For each language, we randomly pick up to 1000 words that have only one English word translation in the crowdsourced dictionary to be the test set for that language. On average, there are 967 test source words with a variety of POS per language. Since different language treats grammatical categories such as tense and number differently (for example, unlike English, tenses are not expressed by specific forms of words in Indonesian (\texttt{id}); rather, they are expressed through context), we make our evaluation on all languages in \textbf{\textsc{CrowdTest}} generic by treating a predicted English translation of a foreign word as correct as long as it has the same lemma as the gold English translation. To facilitate further research, we make \textbf{\textsc{CrowdTest}} publicly available in our website. 

\subsubsection{Bilingual Signals for Translation}
\begin{figure}
	\centering
	\includegraphics[scale=0.35]{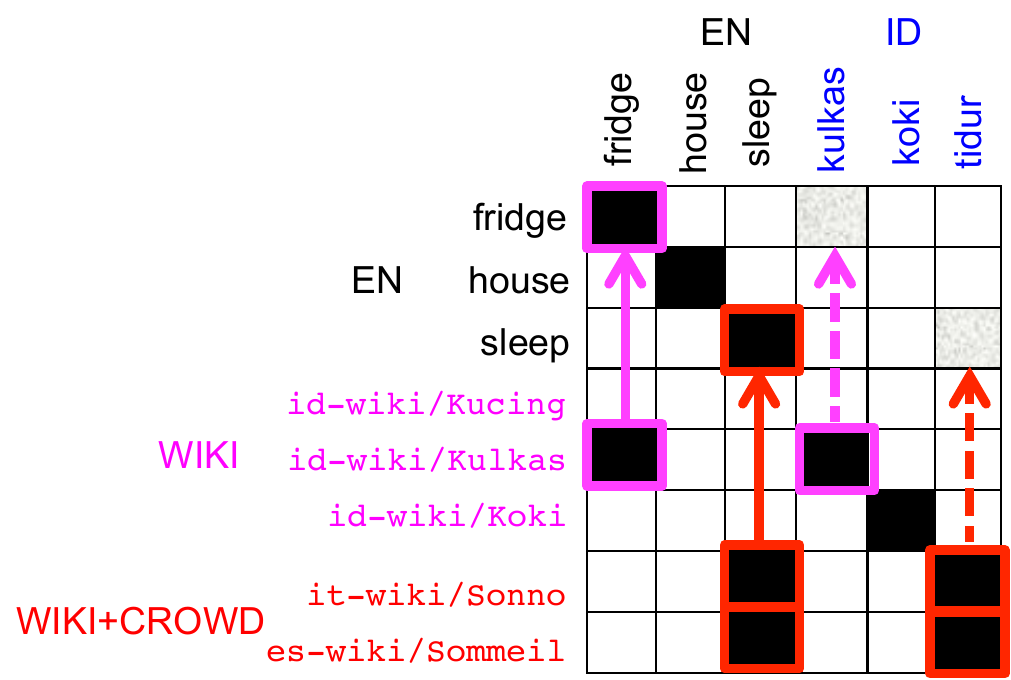}
	\caption{Wikipedia pages with observed translations to the source (id) and the target (en) languages act as a \textit{third} language in the matrix.}
	\label{fig:wiki}
\end{figure}

We use Wikipedia to incorporate information from a \textit{third} language into the matrix, with observed translations to both the source language and the target language, English. We first collect all interlingual links from English Wikipedia pages to pages in other languages. Using these links, we obtain translations of Wikipedia page titles in many languages to English e.g., \texttt{\small id.wikipedia.org/wiki/Kulkas} $\rightarrow$ \textit{fridge} (en). The  observed translations are projected to fill the missing translations in the matrix (Figure \ref{fig:wiki}). We call these bilingual translations \textbf{\texttt{WIKI}}. 

From the links that we have collected, we can also infer links from Wikipedia pages in the source language to other pages in non-target languages e.g., \texttt{\small id.wikipedia.org/wiki/Kulkas} $\rightarrow$ \texttt{\small it.wikipedia.org/wiki/Frigorifero}. The titles of these pages can be translated to English if they exist as entries in the dictionaries.
%\footnote{To obtain high accuracy translations, we use translation entries in crowdsourced dictionaries that have at least 80\% reported accuracy in \citet{pavlick2014language}}. 
These non-source, non-target language pages can act as yet another third language whose observed translations %to the source language and to the target English language 
can be projected to fill the missing translations in the matrix (Figure \ref{fig:wiki}). We call these bilingual translations \textbf{\texttt{WIKI+CROWD}}.

%\begin{figure}
%	\centering
%	\includegraphics[scale=0.35]{wikiandcrowd.pdf}
%	\caption{\textit{Third} language Wikipedia pages with translations to the source (id) and to the target (en) languages obtained by combining Wikipedia interlanguage links and crowdsourced translations.}
%	\label{fig:wikiandcrowd}
%\end{figure}
\subsubsection{Monolingual Signals for Translation}
We define \textit{cold} source words in our experiments as source words that have no associated \textbf{\texttt{WIKI}} translations and fewer than 2 associated \textbf{\texttt{WIKI+CROWD}} translations.
%\footnote{We set a higher threshold for \textbf{\texttt{WIKI+CROWD}} since crowdsourced translations tend to be noisier than Wikipedia interlanguage links}. 
For each cold source word $f$, we predict the score of its translation to each candidate English word $e$ using the auxiliary formulation of $\hat{x}_{e,f}$ (Equation \ref{eq:2}). There are two auxiliary signals about the words that we use in our experiments: (1) bilingually informed word embeddings and (2) visual representations. 

\paragraph{Bilingually Informed Word Embeddings}
For each language, we learn monolingual embeddings for its words by training a standard monolingual \texttt{word2vec} skipgram model \cite{mikolov2013distributed} on tokenized Wikipedia pages of that language using Gensim \cite{rehurek_lrec}. We  obtain 100-dimensional word embeddings with 15 epochs, 15 negatives, window size of 5, and cutoff value of 5.
%(except for Somali, where we use a cutoff of 2)\footnote{For other parameters, we use the default setting of Gensim word2vec script \cite{rehurek_lrec}.}.

Given two monolingual embedding spaces $\mathbb{R}^{d_F}$ and  $\mathbb{R}^{d_E}$ of the source and target languages $F$ and $E$, where $d_f$ and $d_e$ denote the dimensionality of the monolingual embedding spaces, we use the set of crowdsourced translations that are not in the test set %of the source language 
as our seed bilingual translations\footnote{On average, there are 9846 crowdsourced translations per language that we can use as seed translations.} 
and learn a  mapping function $\textbf{W} \in \mathbb{R}^{d_E \times d_F}$ that maps the target language vectors in the seed translations to their corresponding source language vectors.\footnote{We find that mapping from target to source vectors gives better performances across models in our experiments.}

We learn two types of mapping: linear and non-linear, and compare their performances. The linear mapping \cite{mikolov2013exploiting,dinu2014improving} minimizes: $||\mathbf{X}_{\mathbf{E}}\mathbf{W} - \mathbf{X}_{\mathbf{F}} ||^2_F$  where, following the notation in \cite{vulic2016role}, $\mathbf{X}_{\mathbf{E}}$ and $\mathbf{X}_{\mathbf{F}}$ are matrices obtained by respective concatenation of target language and source language vectors that are in the seed bilingual translations. We solve this optimization problem using stochastic gradient descent (SGD). 

We also consider a non-linear mapping ~\cite{socher2013zero} using a simple four-layer neural network, $\textbf{W}$ = ($\phi^{(1)},\phi^{(2)}, \phi^{(3)}, \phi^{(4)}$) that is trained to minimize:

\small
\begin{align*}\sum_{\mathbf{x}_f\in \mathbf{X}_{\mathbf{F}}} \sum_{\mathbf{x}_e \in \mathbf{X}_{\mathbf{E}}} || \mathbf{x}_f - \phi^{(4)} s (\phi^{(3)} s ( \phi^{(2)} s (\phi^{(1)} \mathbf{x}_e)))||^2
\end{align*}\normalsize
where $\phi^{(1)} \in \mathbb{R}^{h_1 \times d_E}$, $\phi^{(2)} \in \mathbb{R}^{h_2 \times h_1}$, $\phi^{(3)} \in \mathbb{R}^{h_3 \times h_2}$, $\phi^{(4)} \in \mathbb{R}^{d_F \times h_3}$, $h_n$ is the size of the hidden layer, and $s = \mathrm{tanh}$ is the chosen non-linear function. 

Once the map $\mathbf{W}$ is learned, all candidate target word vectors $\mathbf{x}_e$ can be mapped into the source language embedding space $\mathbb{R}^{d_{F}}$ by computing $\mathbf{x}_e^T\mathbf{W}$. Instead of the raw monolingual word embeddings $\mathbf{x}_e$, we use these bilingually-informed mapped word vectors $\mathbf{x}_e^T\mathbf{W}$ as the auxiliary word features \textbf{$\texttt{WORD-AUX}$} to estimate $\hat{x}^{\mathrm{AUX}}_{e,f}$. 
\paragraph{Visual Representations Pilot Study}
\begin{figure}
	\centering
	\includegraphics[scale=0.35]{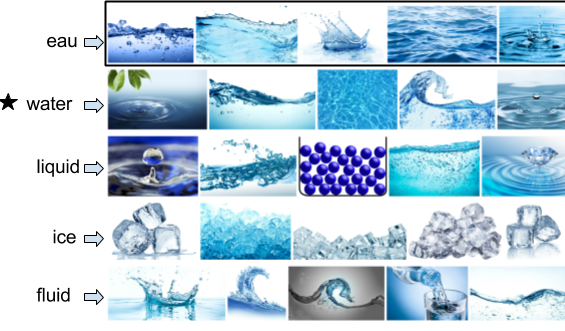}
	\caption{Five images for the French word \textit{eau} and its top 4 translations ranked using visual simularities of images associated with English words \cite{bergsma2011learning}}
	\label{fig:eau}
\end{figure}
Recent work \cite{vulic2016multi} has shown that combining word embeddings and visual representations of words can help achieve more accurate bilingual translations. Since the visual representation of a word seems to be language-independent (e.g. the concept of \textit{water} has similar images whether expressed in English or French (Figure \ref{fig:eau}), the visual representations of a word may be useful for inferring its translation and for complementing the information learned in text. 

We performed a pilot study to include visual features as auxiliary features in our framework.  
We use a large multilingual corpus of labeled images \cite{largemultiimagecorpus} to obtain the visual representation of the words in our source and target languages. The corpus contains 100 images for up to 10k words in each of 100 foreign languages, plus images of each of their translations into English. For each of the images, a convolutional neural network (CNN) feature vector is also provided following the method of \citet{kiela2015visual}. For each word, we use 10 images provided by this corpus and use their CNN features as auxiliary visual features \textbf{$\texttt{VISUAL-AUX}$} to estimate $\hat{x}^{\mathrm{AUX}}_{e,f}$. 

\subsubsection{Combining Signals}
During training, we trained the parameters of $\hat{x}_{e,f}^{\mathrm{MF}}$ and $\hat{x}_{e,f}^{\mathrm{AUX}}$ using a variety of signals: 
\begin{itemize}[noitemsep]
	\item $\hat{x}_{e,f}^{\mathrm{MF}-\mathrm{W}}$ is trained using \textbf{$\texttt{WIKI}$} translations as the set of observed translations $T$
	\item $\hat{x}_{e,f}^{\mathrm{MF}-\mathrm{W+C}}$ is trained using \textbf{$\texttt{WIKI+CROWD}$} translations as the set of observed $T$
	\item $\hat{x}_{e,f}^{\mathrm{AUX}-\mathrm{WE}}$ is trained using the set of word identities $T^{identity}$ and \textbf{$\texttt{WORD-AUX}$} as $\theta_{f}$
	\item $\hat{x}_{e,f}^{\mathrm{AUX}-\mathrm{VIS}}$ is trained using the set of word identities $T^{identity}$ and \textbf{$\texttt{VISUAL-AUX}$} as $\theta_{f}$
\end{itemize}
During testing, we use the following back-off scheme to predict translation scores given a source word $f$ and a candidate target word $e$: 

\small
\[
\hat{x}_{e,f} = 
\begin{cases}
\hat{x}_{e,f}^{\mathrm{MF}-\mathrm{W}}& \text{if } f\ \text{has} \ge 1\ \text{associated\textbf{ \texttt{WIKI}},}\\
\hat{x}_{e,f}^{\mathrm{MF}-\mathrm{W+C}}& \text{else if } f\ \text{has} \ge 2\ \\ & \ \ \ \ \ \ \ \text{associated\textbf{ \texttt{WIKI+CROWD}}},\\
\hat{x}_{e,f}^{\mathrm{AUX}}& \text{otherwise}
\end{cases}
\]
\normalsize
where $\hat{x}_{e,f}^{\mathrm{AUX}} = \alpha_{\mathrm{we}}\  \hat{x}_{e,f}^{\mathrm{AUX}-\mathrm{WE}} + \alpha_{\mathrm{vis}}\  \hat{x}_{e,f}^{\mathrm{AUX}-\mathrm{VIS}}$

\begin{figure*}[ht]
	\centering
	\includegraphics[scale=0.37]{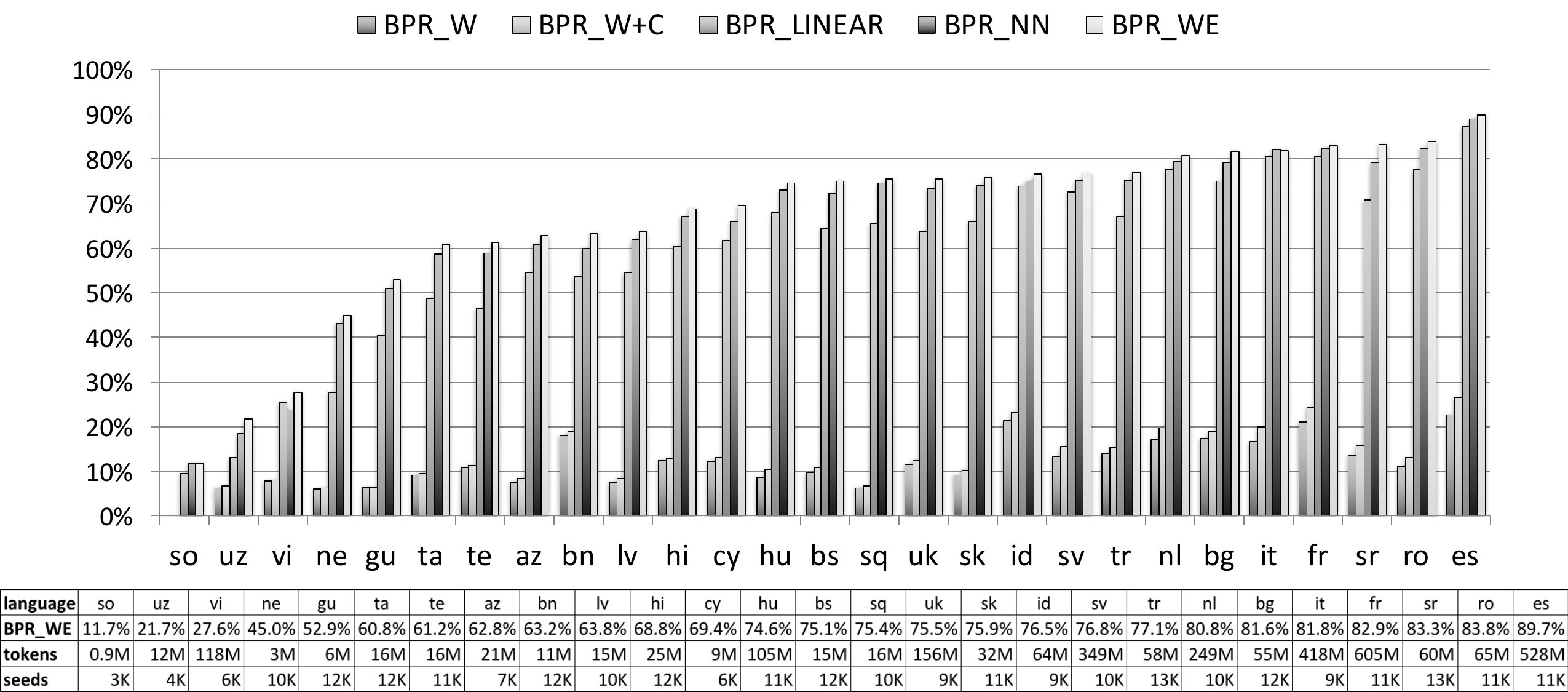}
	\caption{$Acc_{10}$ on \textbf{\textsc{CROWDTest}} across all 27 languages show that adding more and better signals for translation improves translation accuracies. The top accuracies achieved by our model: \textbf{\texttt{BPR\_WE}} vary across languages and appear to be influenced by the amount of data (Wikipedia tokens and seed translations) and tokenization available for the language.  }
	\label{fig:crowdtest}
\end{figure*}

\subsection{Results} 
We conduct experiments using the following variants of our model, each of which progressively incorporates more signals to rank candidate English target words. When a variant uses more than one formulation of $\hat{x}_{e,f}$, it applies them using the back-off scheme that we have described before. 
\begin{itemize}[noitemsep]
	\item \textbf{\texttt{BPR\_W}} uses \textit{only} $\hat{x}_{e,f}^{\mathrm{MF}-\mathrm{W}}$
	\item \textbf{\texttt{BPR\_W+C}} uses $\hat{x}_{e,f}^{\mathrm{MF}-\mathrm{W}}$ and $\hat{x}_{e,f}^{\mathrm{MF}-\mathrm{W+C}}$ 
	\item \textbf{\texttt{BPR\_LN}} uses \textit{only} $\hat{x}_{e,f}^{\mathrm{AUX}-\mathrm{WE}}$ with linear mapping
	\item \textbf{\texttt{BPR\_NN}} uses \textit{only} $\hat{x}_{e,f}^{\mathrm{AUX}-\mathrm{WE}}$ with neural network (NN) mapping
	\item \textbf{\texttt{BPR\_WE}} uses $\hat{x}_{e,f}^{\mathrm{MF}-\mathrm{W}}$, $\hat{x}_{e,f}^{\mathrm{MF}-\mathrm{W+C}}$, and $\hat{x}_{e,f}^{\mathrm{AUX}-\mathrm{WE}}$ with NN mapping
	\item \textbf{\texttt{BPR\_VIS}} adds $\hat{x}_{e,f}^{\mathrm{AUX}-\mathrm{VIS}}$ to \textbf{\texttt{BPR\_WE}}
\end{itemize}
\begin{table}[h]
	\small
	\begin{center}
		\begin{tabular}{|c|c|c|c|c|}
			\hline 
			& \textbf{\texttt{Baseline}}& \textbf{\texttt{BPR+MNN}} & \textbf{\texttt{BPR\_LN}} & \textbf{\texttt{BPR\_WE}} \\ 
			& (\textbf{\texttt{MNN}})&  &  &  \\ 
			\hline 
			IT-EN & 78.8\% & 79.4\% & 81.3\% & 86.0\% \\ 
			\hline 
			ES-EN & 81.8\% & 82.1\% & 83.4\% & 87.1\%\\ 
			\hline 
			NL-EN & 80.8\% & 81.6\% & 83.2\% & 87.2\%\\ 
			\hline 
		\end{tabular} 
		\caption{$Acc_{10}$ performance on \textbf{\textsc{VULIC1000}}}
		\label{table:1}
	\end{center}
	\normalsize
\end{table}
\begin{table*}
	\small
	\begin{center}
		\begin{tabular}{|c|c|c|c|c|}
			\hline 
			\textbf{\texttt{BPR\_W}} & \textbf{\texttt{BPR\_LN}} & \textbf{\texttt{BPR\_NN}} & \textbf{\texttt{BPR\_WE}} \\ 
			\hline 
			\textit{kesadaran} &\textit{kesadaran} & \textit{kesadaran} & \textit{kesadaran} \\
		    \hline 
			consciousness & consciousness & consciousness &  conscience\\
			goddess & \textit{awareness} & empathy & \textit{awareness} \\
			friendship & empathy & \textit{awareness} & understanding \\
			night & perception & perceptions & consciousness \\
			nation & mindedness & perception &  acquaintance\\
			\hline 
		\end{tabular} 
		\caption{Top-5 translations of the Indonesian word \textit{kesadaran} (\textit{awareness}) using different model variants}
		\label{table:0}
	\end{center}
	\normalsize
\end{table*}
We evaluate the performance of \textbf{\texttt{BPR\_WE}} against a baseline that is the state-of-the-art model of \citet{vulic2016role}, on benchmark \textbf{\textsc{VULIC1000}} (Table \ref{table:1}). The baseline (\textbf{\texttt{MNN}}) learns a linear mapping between monolingual embedding spaces and finds translations in an \textit{unsupervised} manner: it ranks candidate target words based on their cosine similarities to the source word in the mapped space. As seed translation pairs, \textbf{\texttt{MNN}} uses mutual nearest neighbor pairs (MNN) obtained from \textit{pseudo-bilingual} corpora constructed from  unannotated monolingual data of the source and target languages~\cite{vulic2016bilingual}. We train \textbf{\texttt{MNN}} and our models using the same 100-dimensional \texttt{word2vec} monolingual word embeddings. %To train \textbf{\texttt{HYBWE}}, we learn a linear mapping (with SGD) between the monolingual embedding spaces using, as seed translation pairs, all mutual nearest neighbor pairs that we discover in the pseudo bilingual space that we construct following \citet{vulic2016bilingual}. 

As seen in Table \ref{table:1}, we see the benefit of learning translations in a supervised manner. \textbf{\texttt{BPR+MNN}} uses the same MNN seed translations as \textbf{\texttt{MNN}}, obtained from unannotated monolingual data of English and the foreign language, to learn the linear mapping between their embedding spaces. However, unlike \textbf{\texttt{MNN}}, \textbf{\texttt{BPR+MNN}} uses the mapped word vectors to predict ranking in a \textit{supervised} manner with BPR objective. This results in higher accuracies than \textbf{\texttt{MNN}}. Using seed translations from crowdsourced dictionaries to learn the linear mapping (\textbf{\texttt{BPR\_LN}}) improves accuracies even further compared to using MNN seed translations obtained from unannotated data. %Learning the mapping non-linearly improves the accuracies even further. 
Finally, \textbf{\texttt{BPR\_WE}} that learns translations in a supervised manner and uses third language translations and non-linear mapping (trained with crowdsourced translations not in the test set) performs consistently and very significantly better than the state-of-the-art on all benchmark test sets. This shows that incorporating more and better signals of translation % e.g., crowdsourced translations, non-linear mapping, 
can improve performance significantly.
 
Evaluating on \textbf{\textsc{CROWDTest}}, we observe a similar trend over all 27 languages (Figure \ref{fig:crowdtest}). Particularly, we see that \textbf{\texttt{BPR\_W}} and \textbf{\texttt{BPR\_W+C}} suffer from the \textit{cold} start issue where there are too few or no observed translations in the matrix to make accurate predictions. Incorporating auxiliary information in the form of bilingually-informed word embeddings improves the accuracy of the predictions dramatically. For many languages, learning these bilingually-informed word embeddings with non-linear mapping improves accuracy even more. The top accuracy scores achieved by the model vary across languages and seem to be influenced by the amount of data i.e., Wikipedia tokens and seed lexicons entries available for training. Somali (\texttt{so}) for example, has only 0.9 million tokens available in its Wikipedia for training the \texttt{word2vec} embeddings and only 3 thousand seed translations for learning the mapping between the word embedding spaces. In comparison, Spanish (\texttt{es}) has over 500 million tokens available in its Wikipedia and 11 thousand seed translations. We also believe that our choice of tokenization may not be suitable for some languages -- we use a simple regular-expression based tokenizer for many languages that do not have a trained \textsc{NLTK}\footnote{http://www.nltk.org/} tokenization model. This may influence performance on languages such as Vietnamese (\texttt{vi}) on which we have a low performance despite its large Wikipedia corpus. 

Some example translations of an Indonesian word produced by different variants of our model are shown in Table \ref{table:0}. Adding third language translation signals on top of the bilingually-informed auxiliary signals improves accuracies even further.\footnote{Actual improvement per language depends on the coverage of the Wikipedia interlanguage links for that language} The accuracies achieved by \textbf{\texttt{BPR\_WE}} on these languages are significantly better than previously reported accuracies \cite{irvine2017comprehensive} on test sets constructed from the same crowdsourced dictionaries \cite{pavlick2014language}\footnote{The comparison however, cannot be made apples-to-apples since the way \citet{irvine2017comprehensive} select test sets from the crowdsourced dictionaries maybe different and they do not release the test sets}. 

The accuracies across languages appear to improve consistently with the amount of signals being input to the model. In the following experiments, we investigate how sensitive these improvements are with varying training size. 

\begin{figure}
	\centering
	\includegraphics[scale=0.27]{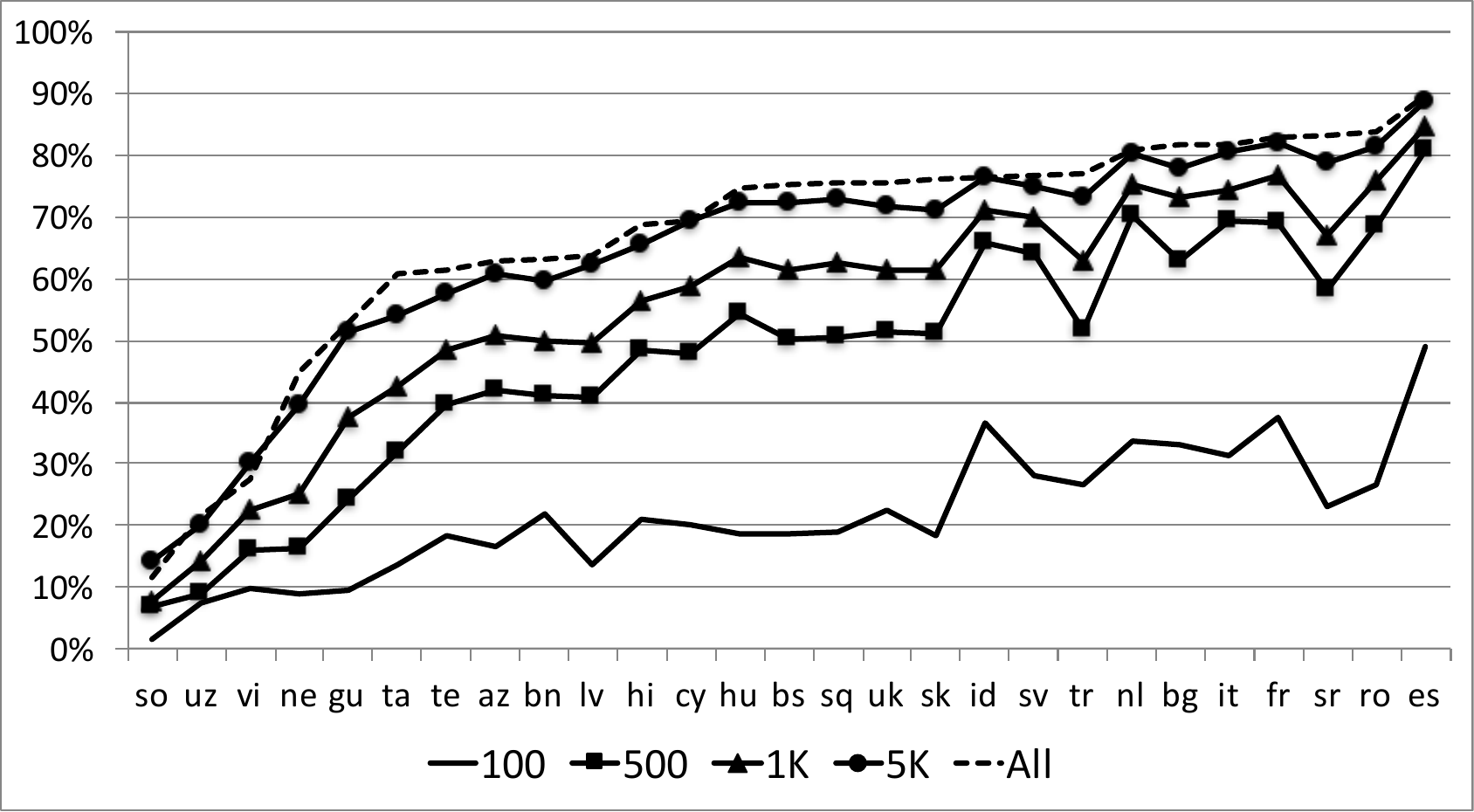}
	\caption{$Acc_{10}$ across different seed lexicon sizes}
	\label{fig:ablation}
\end{figure}

In Figure \ref{fig:ablation}, we show accuracies obtained by \textbf{\texttt{BPR\_WE}} with varying sizes of seed translation lexicons used to train its mapping. The results show that a seed lexicon size of 5K is enough across languages to achieve optimum performance. This finding is consistent with the finding of \citet{vulic2016role} that accuracies peak at about 5K seed translations across all their models and languages. For future work, it will be interesting to investigate further why this is the case: e.g., how optimal seed size is related to the quality of the seed translations and the size of the test set, and how the optimum seed size should be chosen. 
Lastly, we experiment with incorporating auxiliary visual signals for learning translations on the multilingual image corpus \cite{largemultiimagecorpus}. The corpus contains 100 images for up to 10K words in each of 100 foreign languages, plus images of each of their translations into English. We train and test our \textbf{\texttt{BPR\_VIS}} model to learn translations of 5 low- and high-resource languages in this corpus. We use the translations of up to 10K words in each of these languages as test set and use up to 10 images (CNN features) of the words in this set as auxiliary visual signals to predict their translations. In this experiment, we weigh auxiliary word embedding and visual features equally. To train the mapping of our word embedding features, we use as seeds crowdsourced translations not in test set.

We compare the quality of our translations with the baseline CNN-\textsc{AvgMax} \cite{bergsma2011learning}, which considers cosine similarities between individual images from the source and target word languages and takes average of their maximum similarities as the final similarity between a source and a target word. For each source word, the candidate target words are ranked according to these final similarities. This baseline has been shown to be effective for inducing translations from images, both in the uni-modal \cite{bergsma2011learning,kiela2015visual} and multi-modal models \cite{vulic2016multi}. 

As seen in Table \ref{table:2}, incorporating additional bilingual and textual signals to the visual signals improves translations. Accuracies on these image corpus' test sets are lower overall as they contain a lot of translations from our crowdsourced dictionaries; thus we have much less seeds to train our  word embedding mapping. Furthermore, these test sets contain 10 times as many translations as our previous test sets. Using more images instead of just 10 per word may also improve performance. 
\begin{table}
	\small
	\begin{center}
		\begin{tabular}{|c|c|c|r|}
			\hline 
			& \textbf{\texttt{Baseline}} & \textbf{\texttt{BPR\_VIS}}& \# Seeds \\ 
			& (\textbf{\texttt{CNN-AvgMax}}) & & \\ 
			\hline 
			IT-EN & 31.4\% & 55.8\% & 581\\ 
			\hline 
			ES-EN & 33.0\% & 58.3\% & 488\\ 
			\hline 
			NL-EN & 35.5\% & 69.2\% & 1857\\ 
			\hline 
			FR-EN & 37.1\% & 65.9\% & 1697\\ 
			\hline 
			ID-EN & 36.9\% & 45.3\% & 462\\ 
			\hline 
		\end{tabular} 
		\caption{$Acc_{10}$ performance on the multilingual image corpus test set \cite{largemultiimagecorpus}}
		\label{table:2}
	\end{center}
	\normalsize
\end{table}
\section{Conclusion}
In this paper, we propose a novel framework for combining diverse, sparse and potentially noisy multi-modal signals for translations. We view the problem of learning translations as a matrix completion task and use an effective and extendable matrix factorization approach with BPR to learn translations. 

We show the effectiveness of our approach in large scale experiments. Starting from minimally-trained monolingual word embeddings, we  consistently and very significantly outperform state-of-the-art approaches by combining these features with other features in a supervised manner using BPR. Since our framework is modular, each input to our prediction can be improved separately to improve the whole system e.g., by learning better word embeddings or a better mapping function to input into the auxiliary component. Our framework is also easily extendable to incorporate more bilingual and auxiliary signals of translation equivalence.  

\section*{Acknowledgments}

This material is based in part on research sponsored by DARPA under grant number HR0011-15-C-0115 (the LORELEI program). The U.S. Government is authorized to reproduce and distribute reprints for Governmental purposes. The views and conclusions contained in this publication are those of the authors and should not be interpreted as representing official policies or endorsements of DARPA and the U.S. Government.

This work was also supported by the French National Research Agency under project ANR-16-CE33-0013, and by Amazon through the Amazon Academic Research Awards (AARA) program.

\bibliography{bpr}

\begin{thebibliography}{47}
\expandafter\ifx\csname natexlab\endcsname\relax\def\natexlab#1{#1}\fi

\bibitem[{Anonymous(2017)}]{largemultiimagecorpus}
Anonymous. 2017.
\newblock A large multilingual corpus for learning translations from images.
\newblock In \emph{Submission}.

\bibitem[{AP et~al.(2014)AP, Lauly, Larochelle, Khapra, Ravindran, Raykar, and
  Saha}]{ap2014autoencoder}
Sarath~Chandar AP, Stanislas Lauly, Hugo Larochelle, Mitesh Khapra, Balaraman
  Ravindran, Vikas~C Raykar, and Amrita Saha. 2014.
\newblock An autoencoder approach to learning bilingual word representations.
\newblock In \emph{Advances in Neural Information Processing Systems}, pages
  1853--1861.

\bibitem[{Bergsma and Van~Durme(2011)}]{bergsma2011learning}
Shane Bergsma and Benjamin Van~Durme. 2011.
\newblock Learning bilingual lexicons using the visual similarity of labeled
  web images.
\newblock In \emph{IJCAI Proceedings-International Joint Conference on
  Artificial Intelligence}, volume~22, page 1764. Citeseer.

\bibitem[{Callison-Burch et~al.(2006)Callison-Burch, Koehn, and
  Osborne}]{callison2006improved}
Chris Callison-Burch, Philipp Koehn, and Miles Osborne. 2006.
\newblock Improved statistical machine translation using paraphrases.
\newblock In \emph{Proceedings of the North American Chapter of the Association
  of Computational Linguistics}, pages 17--24. Association for Computational
  Linguistics.

\bibitem[{Daum\'e and Jagarlamudi(2011)}]{daume11lexicaladapt}
Hal Daum\'e and Jagadeesh Jagarlamudi. 2011.
\newblock Domain adaptation for machine translation by mining unseen words.
\newblock In \emph{Proceedings of the 49th Annual Meeting of the Association
  for Computational Linguistics}, pages 407--412, Portland, Oregon, USA.
  Association for Computational Linguistics.

\bibitem[{Dinu et~al.(2014)Dinu, Lazaridou, and Baroni}]{dinu2014improving}
Georgiana Dinu, Angeliki Lazaridou, and Marco Baroni. 2014.
\newblock Improving zero-shot learning by mitigating the hubness problem.
\newblock In \emph{ICLR Workshop Papers}.

\bibitem[{Fan et~al.(2014)Fan, Zhao, Zhou, Liu, Zheng, and
  Chang}]{fan2014distant}
Miao Fan, Deli Zhao, Qiang Zhou, Zhiyuan Liu, Thomas~Fang Zheng, and Edward~Y
  Chang. 2014.
\newblock Distant supervision for relation extraction with matrix completion.
\newblock In \emph{ACL (1)}, pages 839--849. Citeseer.

\bibitem[{Faruqui and Dyer(2014)}]{faruqui2014improving}
Manaal Faruqui and Chris Dyer. 2014.
\newblock Improving vector space word representations using multilingual
  correlation.
\newblock Association for Computational Linguistics.

\bibitem[{Gantner et~al.(2010)Gantner, Drumond, Freudenthaler, Rendle, and
  Schmidt-Thieme}]{gantner2010learning}
Zeno Gantner, Lucas Drumond, Christoph Freudenthaler, Steffen Rendle, and Lars
  Schmidt-Thieme. 2010.
\newblock Learning attribute-to-feature mappings for cold-start
  recommendations.
\newblock In \emph{Data Mining (ICDM), 2010 IEEE 10th International Conference
  on}, pages 176--185. IEEE.

\bibitem[{Goutte et~al.(2004)Goutte, Yamada, and Gaussier}]{goutte2004aligning}
Cyril Goutte, Kenji Yamada, and Eric Gaussier. 2004.
\newblock Aligning words using matrix factorisation.
\newblock In \emph{Proceedings of the 42nd Annual Meeting on Association for
  Computational Linguistics}, page 502. Association for Computational
  Linguistics.

\bibitem[{Gouws et~al.(2014)Gouws, Bengio, and Corrado}]{gouws2014bilbowa}
Stephan Gouws, Yoshua Bengio, and Greg Corrado. 2014.
\newblock Bilbowa: Fast bilingual distributed representations without word
  alignments.
\newblock \emph{stat}, 1050:9.

\bibitem[{Gulcehre et~al.(2016)Gulcehre, Ahn, Nallapati, Zhou, and
  Bengio}]{gulcehre2016pointing}
Caglar Gulcehre, Sungjin Ahn, Ramesh Nallapati, Bowen Zhou, and Yoshua Bengio.
  2016.
\newblock Pointing the unknown words.
\newblock In \emph{Proceedings of the 54th Annual Meeting of the Association
  for Computational Linguistics (ACL)}. Association for Computational
  Linguistics.

\bibitem[{He and McAuley(2016)}]{he2016vbpr}
Ruining He and Julian McAuley. 2016.
\newblock {VBPR}: Visual bayesian personalized ranking from implicit feedback.
\newblock In \emph{Thirtieth AAAI Conference on Artificial Intelligence}.

\bibitem[{Irvine and Callison-Burch(2013)}]{irvine2013supervised}
Ann Irvine and Chris Callison-Burch. 2013.
\newblock Supervised bilingual lexicon induction with multiple monolingual
  signals.
\newblock Citeseer.

\bibitem[{Irvine and Callison-Burch(2016)}]{irvine2016end}
Ann Irvine and Chris Callison-Burch. 2016.
\newblock End-to-end statistical machine translation with zero or small
  parallel texts.
\newblock \emph{Natural Language Engineering}, 22(04):517--548.

\bibitem[{Irvine and Callison-Burch(2017)}]{irvine2017comprehensive}
Ann Irvine and Chris Callison-Burch. 2017.
\newblock A comprehensive analysis of bilingual lexicon induction.
\newblock \emph{Computational Linguistics}.

\bibitem[{Kiela et~al.(2015)Kiela, Vulic, and Clark}]{kiela2015visual}
Douwe Kiela, Ivan Vulic, and Stephen Clark. 2015.
\newblock Visual bilingual lexicon induction with transferred convnet features.
\newblock In \emph{Proceedings of the 2015 Conference on Empirical Methods in
  Natural Language Processing (EMNLP 2015)}. ACL.

\bibitem[{Klementiev et~al.(2012{\natexlab{a}})Klementiev, Irvine,
  Callison-Burch, and Yarowsky}]{klementiev2012toward}
Alexandre Klementiev, Ann Irvine, Chris Callison-Burch, and David Yarowsky.
  2012{\natexlab{a}}.
\newblock Toward statistical machine translation without parallel corpora.
\newblock In \emph{Proceedings of the 13th Conference of the European Chapter
  of the Association for Computational Linguistics}, pages 130--140.
  Association for Computational Linguistics.

\bibitem[{Klementiev and Roth(2006)}]{klementiev2006weakly}
Alexandre Klementiev and Dan Roth. 2006.
\newblock Weakly supervised named entity transliteration and discovery from
  multilingual comparable corpora.
\newblock In \emph{Proceedings of the 21st International Conference on
  Computational Linguistics and the 44th annual meeting of the Association for
  Computational Linguistics}, pages 817--824. Association for Computational
  Linguistics.

\bibitem[{Klementiev et~al.(2012{\natexlab{b}})Klementiev, Titov, and
  Bhattarai}]{klementiev2012inducing}
Alexandre Klementiev, Ivan Titov, and Binod Bhattarai. 2012{\natexlab{b}}.
\newblock Inducing crosslingual distributed representations of words.
\newblock In \emph{COLING}.

\bibitem[{Koehn et~al.(2003)Koehn, Och, and Marcu}]{koehn2003statistical}
Philipp Koehn, Franz~Josef Och, and Daniel Marcu. 2003.
\newblock Statistical phrase-based translation.
\newblock In \emph{Proceedings of the 2003 Conference of the North American
  Chapter of the Association for Computational Linguistics on Human Language
  Technology-Volume 1}, pages 48--54. Association for Computational
  Linguistics.

\bibitem[{Koren et~al.(2009)Koren, Bell, and Volinsky}]{koren2009matrix}
Yehuda Koren, Robert Bell, and Chris Volinsky. 2009.
\newblock Matrix factorization techniques for recommender systems.
\newblock \emph{Computer}, 42(8).

\bibitem[{Lu et~al.(2015)Lu, Wang, Bansal, Gimpel, and Livescu}]{lu2015deep}
Ang Lu, Weiran Wang, Mohit Bansal, Kevin Gimpel, and Karen Livescu. 2015.
\newblock Deep multilingual correlation for improved word embeddings.
\newblock In \emph{HLT-NAACL}.

\bibitem[{Luong et~al.(2015)Luong, Pham, and Manning}]{luong2015bilingual}
Thang Luong, Hieu Pham, and Christopher~D Manning. 2015.
\newblock Bilingual word representations with monolingual quality in mind.
\newblock In \emph{Proceedings of the 1st Workshop on Vector Space Modeling for
  Natural Language Processing}, pages 151--159.

\bibitem[{Mikolov et~al.(2013{\natexlab{a}})Mikolov, Le, and
  Sutskever}]{mikolov2013exploiting}
Tomas Mikolov, Quoc~V Le, and Ilya Sutskever. 2013{\natexlab{a}}.
\newblock Exploiting similarities among languages for machine translation.
\newblock \emph{arXiv preprint arXiv:1309.4168}.

\bibitem[{Mikolov et~al.(2013{\natexlab{b}})Mikolov, Sutskever, Chen, Corrado,
  and Dean}]{mikolov2013distributed}
Tomas Mikolov, Ilya Sutskever, Kai Chen, Greg~S Corrado, and Jeff Dean.
  2013{\natexlab{b}}.
\newblock Distributed representations of words and phrases and their
  compositionality.
\newblock In \emph{Advances in neural information processing systems}, pages
  3111--3119.

\bibitem[{Pavlick et~al.(2014)Pavlick, Post, Irvine, Kachaev, and
  Callison-Burch}]{pavlick2014language}
Ellie Pavlick, Matt Post, Ann Irvine, Dmitry Kachaev, and Chris Callison-Burch.
  2014.
\newblock The language demographics of {Amazon Mechanical Turk}.
\newblock \emph{Transactions of the Association for Computational Linguistics},
  2:79--92.

\bibitem[{Rapp(1995)}]{rapp1995identifying}
Reinhard Rapp. 1995.
\newblock Identifying word translations in non-parallel texts.
\newblock In \emph{Proceedings of the 33rd annual meeting on Association for
  Computational Linguistics}, pages 320--322. Association for Computational
  Linguistics.

\bibitem[{{\v R}eh{\r u}{\v r}ek and Sojka(2010)}]{rehurek_lrec}
Radim {\v R}eh{\r u}{\v r}ek and Petr Sojka. 2010.
\newblock {Software Framework for Topic Modelling with Large Corpora}.
\newblock In \emph{{Proceedings of the LREC 2010 Workshop on New Challenges for
  NLP Frameworks}}, pages 45--50, Valletta, Malta. ELRA.
\newblock \url{http://is.muni.cz/publication/884893/en}.

\bibitem[{Rendle et~al.(2009)Rendle, Freudenthaler, Gantner, and
  Schmidt-Thieme}]{rendle2009bpr}
Steffen Rendle, Christoph Freudenthaler, Zeno Gantner, and Lars Schmidt-Thieme.
  2009.
\newblock {BPR}: Bayesian personalized ranking from implicit feedback.
\newblock In \emph{Proceedings of the twenty-fifth conference on uncertainty in
  artificial intelligence}, pages 452--461. AUAI Press.

\bibitem[{Riedel et~al.(2013)Riedel, Yao, McCallum, and
  Marlin}]{riedel2013relation}
Sebastian Riedel, Limin Yao, Andrew McCallum, and Benjamin~M Marlin. 2013.
\newblock Relation extraction with matrix factorization and universal schemas.
\newblock In \emph{Proceedings of NAACL-HLT}, pages 74--84.

\bibitem[{Rockt{\"a}schel et~al.(2015)Rockt{\"a}schel, Singh, and
  Riedel}]{rocktaschel2015injecting}
Tim Rockt{\"a}schel, Sameer Singh, and Sebastian Riedel. 2015.
\newblock Injecting logical background knowledge into embeddings for relation
  extraction.
\newblock In \emph{HLT-NAACL}, pages 1119--1129.

\bibitem[{Schafer and Yarowsky(2002)}]{schafer2002inducing}
Charles Schafer and David Yarowsky. 2002.
\newblock Inducing translation lexicons via diverse similarity measures and
  bridge languages.
\newblock In \emph{proceedings of the 6th conference on Natural language
  learning-Volume 20}, pages 1--7. Association for Computational Linguistics.

\bibitem[{Sennrich et~al.(2016{\natexlab{a}})Sennrich, Haddow, and
  Birch}]{sennrich2015improving}
Rico Sennrich, Barry Haddow, and Alexandra Birch. 2016{\natexlab{a}}.
\newblock Improving neural machine translation models with monolingual data.
\newblock In \emph{Proceedings of the 54th Annual Meeting of the Association
  for Computational Linguistics}. Association for Computational Linguistics.

\bibitem[{Sennrich et~al.(2016{\natexlab{b}})Sennrich, Haddow, and
  Birch}]{sennrich2015neural}
Rico Sennrich, Barry Haddow, and Alexandra Birch. 2016{\natexlab{b}}.
\newblock Neural machine translation of rare words with subword units.
\newblock In \emph{Proceedings of the 54th Annual Meeting of the Association
  for Computational Linguistics (ACL)}, pages 1715--1725. Association for
  Computational Linguistics.

\bibitem[{Shi et~al.(2015)Shi, Liu, Liu, and Sun}]{shi2015learning}
Tianze Shi, Zhiyuan Liu, Yang Liu, and Maosong Sun. 2015.
\newblock Learning cross-lingual word embeddings via matrix co-factorization.
\newblock In \emph{ACL (2)}, pages 567--572.

\bibitem[{Socher et~al.(2013)Socher, Ganjoo, Manning, and Ng}]{socher2013zero}
Richard Socher, Milind Ganjoo, Christopher~D Manning, and Andrew Ng. 2013.
\newblock Zero-shot learning through cross-modal transfer.
\newblock In \emph{Advances in neural information processing systems}, pages
  935--943.

\bibitem[{Upadhyay et~al.(2016)Upadhyay, Faruqui, Dyer, and
  Roth}]{bicompare:16}
Shyam Upadhyay, Manaal Faruqui, Chris Dyer, and Dan Roth. 2016.
\newblock Cross-lingual models of word embeddings: An empirical comparison.
\newblock In \emph{Proc. of ACL}.

\bibitem[{Verga et~al.(2016)Verga, Belanger, Strubell, Roth, and
  McCallum}]{verga2016multilingual}
Patrick Verga, David Belanger, Emma Strubell, Benjamin Roth, and Andrew
  McCallum. 2016.
\newblock Multilingual relation extraction using compositional universal
  schema.
\newblock In \emph{Proceedings of NAACL-HLT}, pages 886--896.

\bibitem[{Vuli{\'c} et~al.(2016)Vuli{\'c}, Kiela, Clark, and
  Moens}]{vulic2016multi}
Ivan Vuli{\'c}, Douwe Kiela, Stephen Clark, and Marie-Francine Moens. 2016.
\newblock Multi-modal representations for improved bilingual lexicon learning.
\newblock In \emph{Proceedings of the 54th Annual Meeting of the Association
  for Computational Linguistics}, pages 188--194. ACL.

\bibitem[{Vuli{\'c} and Korhonen(2016)}]{vulic2016role}
Ivan Vuli{\'c} and Anna Korhonen. 2016.
\newblock On the role of seed lexicons in learning bilingual word embeddings.
\newblock In \emph{Proceedings of the 54th Annual Meeting of the Association
  for Computational Linguistics (ACL)}. Association for Computational
  Linguistics.

\bibitem[{Vuli{\'c} and Moens(2013)}]{vulic2013study}
Ivan Vuli{\'c} and Marie-Francine Moens. 2013.
\newblock A study on bootstrapping bilingual vector spaces from non-parallel
  data (and nothing else).
\newblock In \emph{Proceedings of the 2013 Conference on Empirical Methods in
  Natural Language Processing (EMNLP 2013)}, pages 1613--1624. ACL.

\bibitem[{Vuli{\'c} and Moens(2015)}]{vulic2015bilingual}
Ivan Vuli{\'c} and Marie-Francine Moens. 2015.
\newblock Bilingual word embeddings from non-parallel document-aligned data
  applied to bilingual lexicon induction.
\newblock In \emph{Proceedings of the 53rd Annual Meeting of the Association
  for Computational Linguistics (ACL 2015)}, pages 719--725. ACL.

\bibitem[{Vuli{\'c} and Moens(2016)}]{vulic2016bilingual}
Ivan Vuli{\'c} and Marie-Francine Moens. 2016.
\newblock Bilingual distributed word representations from document-aligned
  comparable data.
\newblock \emph{The Journal of Artificial Intelligence Research}.

\bibitem[{Wu et~al.(2016)Wu, Schuster, Chen, Le, Norouzi, Macherey, Krikun,
  Cao, Gao, Macherey et~al.}]{wu2016google}
Yonghui Wu, Mike Schuster, Zhifeng Chen, Quoc~V Le, Mohammad Norouzi, Wolfgang
  Macherey, Maxim Krikun, Yuan Cao, Qin Gao, Klaus Macherey, et~al. 2016.
\newblock Google's neural machine translation system: Bridging the gap between
  human and machine translation.
\newblock \emph{arXiv preprint arXiv:1609.08144}.

\bibitem[{Zhou et~al.(2013)Zhou, Liu, Liu, He, Zhao
  et~al.}]{zhou2013statistical}
Guangyou Zhou, Fang Liu, Yang Liu, Shizhu He, Jun Zhao, et~al. 2013.
\newblock Statistical machine translation improves question retrieval in
  community question answering via matrix factorization.
\newblock In \emph{ACL (1)}, pages 852--861.

\bibitem[{Zou et~al.(2013)Zou, Socher, Cer, and Manning}]{zou2013bilingual}
Will~Y Zou, Richard Socher, Daniel~M Cer, and Christopher~D Manning. 2013.
\newblock Bilingual word embeddings for phrase-based machine translation.
\newblock In \emph{EMNLP}, pages 1393--1398.

\end{thebibliography}


\begin{thebibliography}{4}
\expandafter\ifx\csname natexlab\endcsname\relax\def\natexlab#1{#1}\fi

\bibitem[{Aho and Ullman(1972)}]{Aho:72}
Alfred~V. Aho and Jeffrey~D. Ullman. 1972.
\newblock \emph{The Theory of Parsing, Translation and Compiling}, volume~1.
\newblock Prentice-Hall, Englewood Cliffs, NJ.

\bibitem[{{American Psychological Association}(1983)}]{APA:83}
{American Psychological Association}. 1983.
\newblock \emph{Publications Manual}.
\newblock American Psychological Association, Washington, DC.

\bibitem[{Chandra et~al.(1981)Chandra, Kozen, and Stockmeyer}]{Chandra:81}
Ashok~K. Chandra, Dexter~C. Kozen, and Larry~J. Stockmeyer. 1981.
\newblock \href {https://doi.org/10.1145/322234.322243} {Alternation}.
\newblock \emph{Journal of the Association for Computing Machinery},
  28(1):114--133.

\bibitem[{Gusfield(1997)}]{Gusfield:97}
Dan Gusfield. 1997.
\newblock \emph{Algorithms on Strings, Trees and Sequences}.
\newblock Cambridge University Press, Cambridge, UK.

\end{thebibliography}
\bibliographystyle{emnlp_natbib}

\end{document}